\documentclass{article} 
\usepackage{iclr2020_conference,times}

\usepackage{amsmath,amsfonts,bm}

\def\eqref#1{equation~\ref{#1}}

\def\1{\bm{1}}

\DeclareMathAlphabet{\mathsfit}{\encodingdefault}{\sfdefault}{m}{sl}
\SetMathAlphabet{\mathsfit}{bold}{\encodingdefault}{\sfdefault}{bx}{n}

\usepackage{hyperref}
\usepackage{url}

\usepackage{natbib}
\usepackage{subfigure}
\usepackage{graphicx}
\usepackage{amsmath}
\usepackage{multirow}
\usepackage{verbatim}
\usepackage{booktabs}
\let\sss = \scriptscriptstyle

\title{The Detection of Distributional Discrepancy for Text Generation}

\author{Xingyuan Chen \\
Dept. of Computer Science and Technology \\
Nanjing University\\
Nanjing, China \\
\texttt{1045258214@qq.com} \\
\And
Ping Cai \\
School of Information Science and technology\\
Southwest Jiaotong University \\
Chengdu, China \\
\texttt{1061185275@qq.com} \\
\And
Peng Jin\thanks{Peng Jin, Xingyu  Dai and Jiajun Chen are the co-corresponding authors.} \\
School of Computer Science \\
Leshan Normal University \\
Leshan, China \\
\texttt{jandp@pku.edu.cn} \\
\And
Haokun Du \& Hongjun Wang \\
School of Information Science and technology\\
Southwest Jiaotong University \\
Chengdu, China \\
\texttt{wanghongjun@swjtu.edu.cn} \\
\And
Haokun Du \& Hongjun Wang \\
School of Information Science and technology\\
Southwest Jiaotong University \\
Chengdu, China \\
\texttt{\{daixinyu,chenjj\}@nju.edu.cn} \\
}

\iclrfinalcopy 
\begin{document}

\maketitle

\begin{abstract}
The text generated by neural language models is not as good as the real text. This means that their distributions are different. Generative Adversarial Nets (GAN) are used to alleviate it. However, some researchers argue that GAN variants do not work at all. When both sample quality (such as Bleu) and sample diversity (such as self-Bleu) are taken into account, the GAN variants even are worse than a well-adjusted language model. But, Bleu and self-Bleu can not precisely measure this distributional discrepancy. In fact, how to measure the distributional discrepancy between real text and generated text is still an open problem. In this paper, we theoretically propose two metric functions to measure the distributional difference between real text and generated text. Besides that, a method is put forward to estimate them. First, we evaluate language model with these two functions and find the difference is huge. Then, we try several methods to use the detected discrepancy signal to improve the generator. However the difference becomes even bigger than before. Experimenting on two existing language GANs, the distributional discrepancy between real text and generated text increases with more adversarial learning rounds. It demonstrates both of these language GANs fail. 
\end{abstract}

\section{Introduction}
Text generation by neural language models (LM), such as LSTM \citep{hochreiter1997long} have given rise to much progress and are now used to dialogue generation \citep{Li2017Adversarial}, machine translation \citep{Wu2016Google} and image caption \citep{xu2015show}. However, the generated sentences are still low quality as regards semantics and global coherence and not even perfect grammatically speaking \citep{Caccia2019falling}. 

These issues give rise to large discrepancy between generated text and real text. One underlying reason is the architecture and number of parameters of the LM itself \citep{Radford19,santoro2018relational}. Many researchers attribute this to exposure bias \citep{Bengio2015Scheduled} because the LM is trained with a maximum likelihood estimate (MLE) and predicts the next word conditioned on words from the ground-truth during training. Yet it only conditions on words generated by itself during reference.  

Statistically, this discrepancy means the two distributional functions of real texts and generated texts is different.  Reducing this distributional difference may be a practicable way to improve text generation. 

Some researchers try to reduce this difference with GAN \citep{Goodfellow2014Generative}. They use a discriminator to detect the discrepancy between real samples and generated samples, and feed the signal back to upgrade the generator (a LM). In order to solve the non-differential issue that arises by the need to handle discrete tokens,  reinforcement learning (RL)  \citep{williams1992simple} is adapted by SeqGAN \citep{Yu2016SeqGAN}, RankGAN \citep{Lin2017Adversarial}, and LeakGAN \citep{Guo2017Long}. The Gumble-Softmax is also introduced by GSGAN \citep{Jang2016Categorical} and RelGAN \citep{Nie2019ICLR} to solve this issue. These language GANs pre-train both the generator ($G$) and the discriminator ($D$) before adversarial learning\footnote{An exception is RelGAN which does not need not to pre-train $D$.}. During adversarial learning, for each round, the $G$ is trained several epochs and then, the $D$ is trained tens of epochs. Learning stops when the model converges. Furthermore, considering the generated texts' quality and diversity simultaneously \citep{Shi2018Toward}, MaskGAN \citep{Fedus2018MaskGAN}, DpGAN \citep{XuDiversity}, FMGAN \citep{chen2018adversarial} and RelGAN \citep{Nie2019ICLR} are proposed. They evaluate the generated text with  Bleu and self-Bleu \citep{Zhu2018Texygen} or LM score and reverse LM score \citep{DBLP:journals/corr/abs-1804-07972}, and claim these GANs improve  the performance of the generator.
 
However recently questions have been raised over these claims. \cite{Semeniuta2018On} and \cite{Caccia2019falling} showed that via more precise experiments and evaluation, these considered GAN variants are out-performed by a well-adjusted language model . \cite{dAutu2019Scratch} trained language GANs from scratch, nevertheless, they only achieve "comparable" performance against the LM. \cite{He2019Quantifying} quantifies the exposure bias and concludes it is either 3 percent lower in performance or indistinguishable. 

All the aforementioned methods treat GAN as a black box for evaluation. For those language GANs, there are several critical issues which have not been resolved, such as whether the $D$ detects the discrepancy, whether the detected discrepancy is severe, and whether the signals from $D$ can improve the generator. In this paper, we try to solve these problems via investigating GAN in both pre-training and the adversarial learning process. Theoretically analysing the signal from $D$, we obtain two metric functions to measure the distributional difference. With these two functions, we first measure the difference between the real text and the generated text by a MLE-trained language model (pre-train). Second, we try some methods to update the generator with a feedback signal from $D$, then, we use these metric functions to evaluate the updated generator. Finally, we analysis the existing language GANs during adversarial learning with these two functions. All the code and data can be find https://github.com/.

Our contributions are as follows:

\begin{itemize}
\item We propose two metric functions to measure the distributional difference between real text and generated text. In addition, a method is put forward to estimate these.

\item Evaluated using these two functions, a number of experiments show there is an obvious discrepancy between the real text and the generated text even when it is generated by a well-adjusted language model.

\item Although this discrepancy could be detected by $D$, the feedback signal from $D$ can not improve $G$ using  existing methods. 

\item Experimenting on two existing language GANs, SeqGAN and RelGAN, the distributional discrepancy between real text and generated text increases with more adversarial learning rounds. This demonstrates that both of these language GANs fail. 
\end{itemize}

\section{Method}
In GAN, the generator $G_{\theta}$ implicitly defines a probability distribution $p_{\sss\theta}(x)$ to mimic the real data distribution $p_{\sss d}(x)$.

\begin{equation}\label{equ1}
\mathop{min}\limits_{G_\theta}\mathop{max}\limits_{D_\phi}V(D_\phi, G_\theta) = \mathbb{E}_{x\sim{p_{d}(x)}}\bigg[{logD_\phi(x)}\bigg]+\mathbb{E}_{x_\sim{p_\theta(x)}}\bigg[{log\big(1 - D_\phi(x)\big)}\bigg]
\end{equation}

We define $D_\phi$ to detect the discrepancy between $p_\theta(x)$ and $p_{d}(x)$. We optimize $D_\phi$ as follows:

\begin{equation}\label{equD}
\mathop{max}\limits_{D_\phi}V(D_\phi, G_\theta) = \mathop{max}\limits_{D_\phi} \mathbb{E}_{x\sim{p_{d}}}\bigg[{logD_\phi(x)}\bigg]+\mathbb{E}_{x_\sim{p_\theta}}\bigg[{log\big(1 - D_\phi(x)\big)}\bigg]
\end{equation}

Assuming $D_{\phi}^{*}(x)$ is the optimal solution for a given $\theta$, according to \citep{Goodfellow2014Generative}, it will be,

\begin{equation}\label{equBestD}
D_{\phi}^{*}(x) = \frac{p_{d}(x)}{p_{d}(x) + p_\theta(x)}
\end{equation}

We obtain two metric functions to measure this discrepancy. This sentence should be put the next paragraph.

\begin{equation}\label{Equ4}
    \begin{cases}
    D_{\phi}^{*}(x) \geq 0.5,   \qquad iif \quad p_d(x) \geq p_\theta(x)\\[2ex]
    D_{\phi}^{*}(x) < 0.5,   \qquad iif \quad p_d(x) < p_\theta(x)
    \end{cases}
\end{equation}

With this, the integration of the density function can be transformed into an equation for a statistic. Based on this, we can obtain a way which, described below, to compute the precise discrepancy. 

\subsection{Approximate Discrepancy}
Let,

\begin{equation}\label{Equ5-2}
    q_d(x) = \frac{p_d(x)}{p_d(x) + p_\theta(x)}  \\
    \qquad q_\theta(x) = \frac{p_\theta(x)}{p_d(x) + p_\theta(x)}
\end{equation}

\begin{figure*}[htbp]
\centering
\subfigure[Approximate discrepancy illustration.]{
\includegraphics[width=6.5cm]{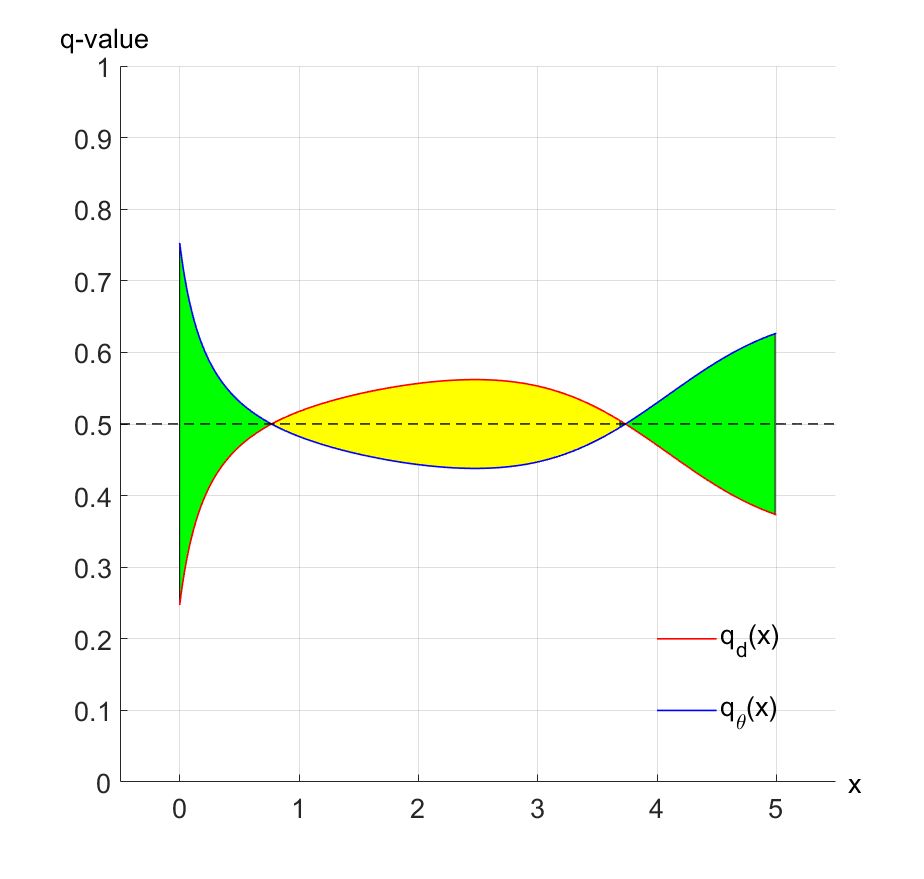}
}
\quad
\subfigure[Precise discrepancy illustration.]{
\includegraphics[width=6.5cm]{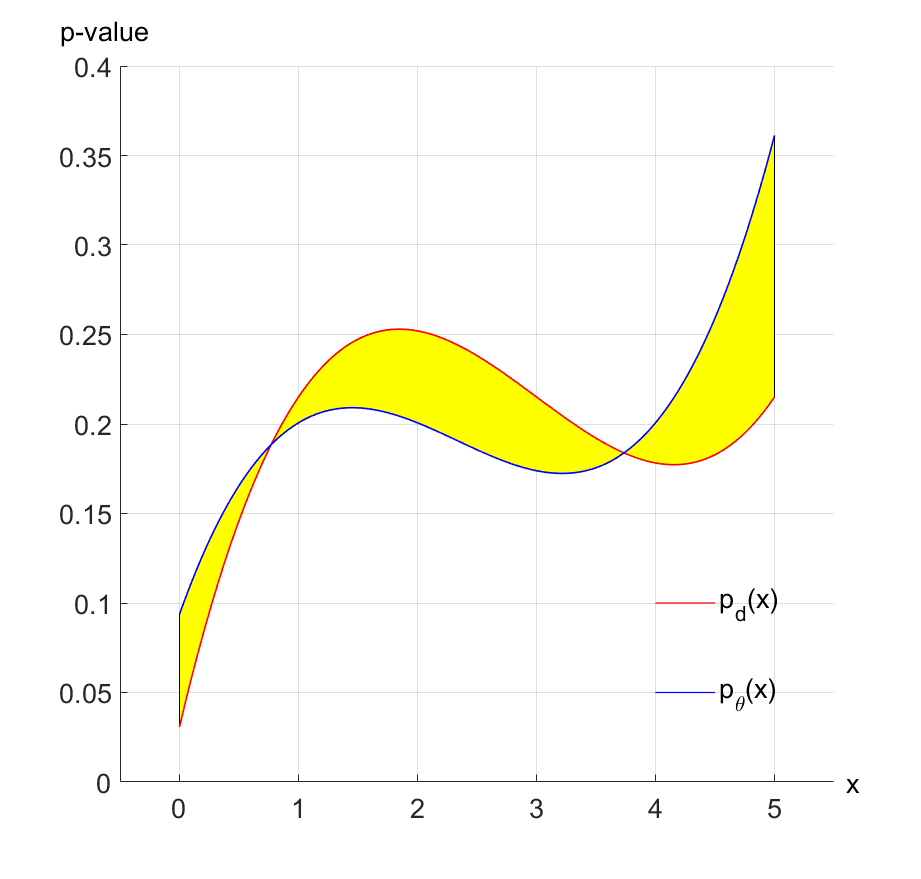}
}
\caption{The illustration of two measures. In (a), the yellow area denotes the negative and the green one denotes the positive. In (b), the half of the shaded area equals the result of equation \ref{equ8}. Therefore, the range is $0\sim2$. Larger values indicate more discrepancy.}
\label{figDiscrepancy}
\end{figure*}

So, $q_d(x) = p(x$ comes from real data$|x)$, $q_\theta(x) = p(x$ comes from generated data$|x)$, $q_d(x) + q_\theta(x) =1$. With equation \ref{Equ5-2}, we can get a constraint and an approximate measure of distributional function. Figure \ref{figDiscrepancy}(a) illustrates the relationship between $q_\theta(x)$ and $q_d(x)$.

Let,
\begin{equation}\label{Equ5}
    u_d = \mathbb{E}_{x\sim{p_{d}(x)}}\big(D_{\phi}^{*}(x)\big)  \\
    \qquad u_\theta = \mathbb{E}_{x\sim{p_{\theta}(x)}}\big(D_{\phi}^{*}(x)\big)
\end{equation}

These are two equations for these two statistics which are the expectation of the $D_{\phi}^{*}$'s predictions on real text and on generated text respectively. From the above equation, it is easy to obtain the following equation: 

\begin{equation}\label{equ6-2}
\frac{1}{2}\big[u_{\sss d} + u_{\sss\theta}\big] = 0.5
\end{equation}

It gives a constraint for $D_\phi$ converging to $D_{\phi}^{*}$. We should take this constraint into account when estimating the ideal function $D_{\phi}^{*}$. From equation \ref{equBestD}, the process of optimizing the discriminator is to make $u_d$ big and make $u_\theta$ small. So, we can estimate the distributional discrepancy according to the following function.

Intuitively, using $u_d$ and $u_\theta$, we get a metric function to measure the discrepancy between $p_\theta(x)$ and $p_{d}(x)$,

\begin{equation}\label{equ6-1}
d_a = u_{\sss d} - u_{\sss\theta} 
\end{equation}

We call this approximate discrepancy. It is the difference in the average score that a well-trained discriminator (denoted as $\hat{D}_\phi$) makes in the predictions on real samples compared to generated samples. It reflects the discrepancy between these two sets to some degree. From equation \ref{Equ5-2},  \ref{Equ5} and \ref{equ6-1}, we get equation \ref{equ6},

\begin{equation}\label{equ6}
d_a  = \int{\big[q_d(x) - q_\theta(x)\big]p_d(x)dx} = \mathbb{E}_{x\sim{p_{d}(x)}}\big[q_d(x) - q_\theta(x)\big]
\end{equation}

Figure \ref{figDiscrepancy} (a) illustrates the discrepancy between two distributional functions $q_\theta(x)$ and $q_{d}(x)$. Both of them are systematic to the line of $q=0.5$. But it is not a complete measure because there is not only a positive part but also a negative part. A complete metric function is given in the next section. 

\subsection{Absolute Discrepancy}
In order to precisely measure the discrepancy, we define $d_s$, 

\begin{equation}\label{equ7}
d_s = \frac{1}{2}\int{\big|p_d(x) - p_\theta(x)\big|}dx
\end{equation}

The range of this function is $0\sim1$. The bigger its value is, the larger the discrepancy. When its value is zero, it means $p_d(x) \equiv p_\theta(x)$, namely there is no discrepancy. Fortunately, this function can be estimated by using a statistic method which is described by the following equation. The proof is shown in appendix A. 

\begin{equation}\label{equ8}
    	d_{s}= \frac{1}{2}\bigg[
    	\mathbb{E}_{\substack{x\sim{p_d(x)}\\ D_{\phi}^{*}(x)>0.5}}\big(1\big) -  
    	\mathbb{E}_{\substack{x\sim{p_d(x)}\\ D_{\phi}^{*}(x)\leq0.5}}\big(1\big) + 
    	\mathbb{E}_{\substack{x\sim{p_\theta(x)}\\ D_{\phi}^{*}(x)\leq0.5}}\big(1\big) - 
    	\mathbb{E}_{\substack{x\sim{p_\theta(x)}\\ D_{\phi}^{*}(x)>0.5}}\big(1\big)
    	\bigg]
\end{equation}

With equation \ref{equ8}, we can more precisely estimate the discrepancy between $p_\theta(x)$ and $p_{d}(x)$. Assuming the classification precision of $D_{\phi}^{*}$ is $a$, then the error rate is $b=1-a$. According to equation \ref{equ8}, $d_{s}=a-b$. So, the discrepancy between $p_\theta(x)$ and $p_{d}(x)$ equals the classification precision of $D_{\phi}^{*}$ minus its error rate.

\subsection{Using $D_{\phi}^{*}(x)$ to improve $G_\theta$}
Given an instance $x$ generated by $G_\theta$, if $D_{\phi}^{*}(x)$ is larger, it means the possibility of $x$ in real data is larger. For an instance $D_{\phi}^{*}(x) = 0.8$, there will be $p_\theta(x) < p_{d}(x)$ according to equation \ref{equBestD}. So, we should update $G_\theta$  to make the probability density $p_\theta(x)$ increase. It may improve the performance of $G_\theta$. Based on this, we can select out some generated instances by the value of $D_{\phi}^{*}(x)$ to update the generator. In fact, we find it helpful to use the faked samples whose score are higher assigned by $D_{\phi}^{*}$. Experiment 4.3 shows the results.

\section{Implementation Procedure}
The optimal function $D_{\phi}^{*}$ is an ideal function which can only be statistically estimated by an approximated function. We can design a function $D_\phi$ and sample from real data and generated data, then train $D_\phi$ according to equation \ref{equD}. When it convergences, we get $\hat{D}_\phi$. $\hat{D}_\phi$ is the approximated function of $D_{\phi}^{*}$. The degree of approximation is mainly determined by three factors: the structure and the number of the parameters number of $D_{\phi}$, the volume of training data, and the settings of hyper-parameters. 

Based on the above analysis, we get two metric functions to measure the distributional discrepancy between dataset $A$ and $B$. The specific implementation procedure is as follows:

\begin{flushleft}
\qquad Step 1: Design a discriminator $D_\phi$.

\qquad Step 2: The sets $A$ and $B$ are respectively each divided into a training set $D_{trainA}$ and $D_{trainB}$, a validation set $D_{devA}$ and $D_{devB}$, and a test set $D_{testA}$ and $D_{testB}$. The partition should be as equal an amount of instances as possible for classification training.

\qquad Step3: $D_\phi$ is optimized with $D_{trainA}$ and $D_{trainB}$ according to the equation \ref{equD}. Validated with $D_{devA}$ and $D_{devB}$, we can judge whether $D_\phi$ convergences or not and then get $\hat{D}_\phi$. 

\qquad Step4: According to equation \ref{equ6-1} and \ref{equ8}, with two test datasets, we can estimate the discrepancy of two distributional density functions between dataset $A$ and $B$. $\hat{d}_s$ denotes the absolute discrepancy and $\hat{d}_a$ denotes the approximate discrepancy respectively. 
\end{flushleft}

Generally speaking, there should be $d_s \leq \hat{d}_s$. Because $D_{\phi}^{*}$ can not be obtained, it is hard to get the degree of the approximation $d_s$ to $\hat{d}_s$. Many research results have shown that the discriminators with deep neural networks are very powerful, and can even exceed human performance on some tasks such as image classification \citep{He2016Residual}  and text classification \citep{Kim2014CNN}. So, if the $D_\phi$ with CNN and attention mechanism is well trained, $\hat{D}_\phi$ will be a meaningful approximation of $D_{\phi}^{*}$. Therefore, we can obtain the meaningful approximation of $d_s$ and $d_a$ via $\hat{D}_\phi$.

\section{Experiment}
We select SeqGAN and RelGAN as representative models for our experiment and the benchmark datasets are also the same as used by these models previously. Then, we show that the well trained discriminator $D$ can measure the discrepancy between the real and generated texts, and then point out the existing GAN-based methods does not work. Finally, a third party discriminator is used to evaluate the performance of adversarial learning with incremental training iterations. 

\subsection{Datasets and Model Settings}
Both SeqGAN and RelGAN are experimented on using relative short sentences (COCO image caption) \footnote{http://cocodataset.org/} and a long sentences dataset (EMNLP2017 WMT news) \footnote{http://www.statmt.org/wmt17/}. For the former dataset, the sentences average
length is about 11 words. There are in total 4,682 word types and the longest sentence consists of 37 words. Both the training and test data contain 10,000 sentences. For the latter dataset, the average length of sentences is about 20 words. There are in total 5,255 word types and the longest sentence consists of 51 words. All training data, about 280 thousand sentences, is used and there are 10,000 sentences in the test data. According to section 3, each test data is divided into two parts. Half is the validation set and the remaining half is the test set. We always generate the same amount of sentences to compare with the two test datasets respectively.

For these two models, all hyper-parameters including word embedding size, learning rate and dropout are set the same as in their original papers. For RelGAN, the standard GAN loss function (the non-saturating version) is adapted because the relative standard loss which is used in \citep{Nie2019ICLR} does not meet the constraints of equation \ref{equ6-2}. But, when measuring RelGAN's discrepancy during the adversarial stage, it own loss function is still relative standard loss. A critical hyper-parameters, temperature, is set to 100 which is the best result in their paper. During the process of training $D$, we always train  10,000 epochs and observe performance on the validation dataset.

\subsection{The distributional difference in Pre-training}
\begin{figure*}[htbp]
\centering
\subfigure[The discrepancy between validation set and generated data. The orange lines denote the absolute discrepancy and the blue lines denote the approximate discrepancy.]{
\includegraphics[width=0.47\columnwidth]{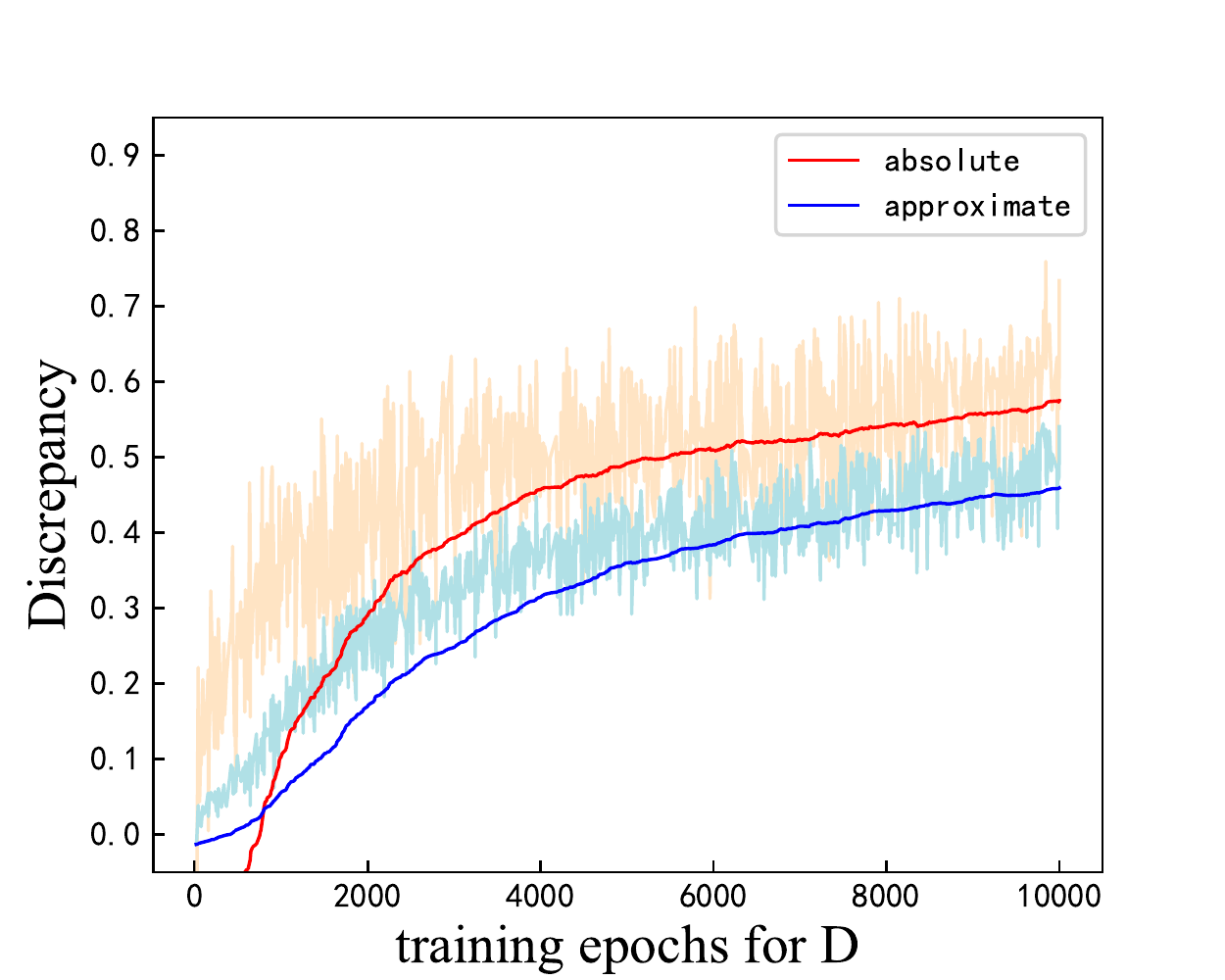}
}
\quad
\subfigure[The discriminator's prediction. The orange lines denote the predictions on validation set and the blue lines denote on generated data.]{
\includegraphics[width=0.47\columnwidth]{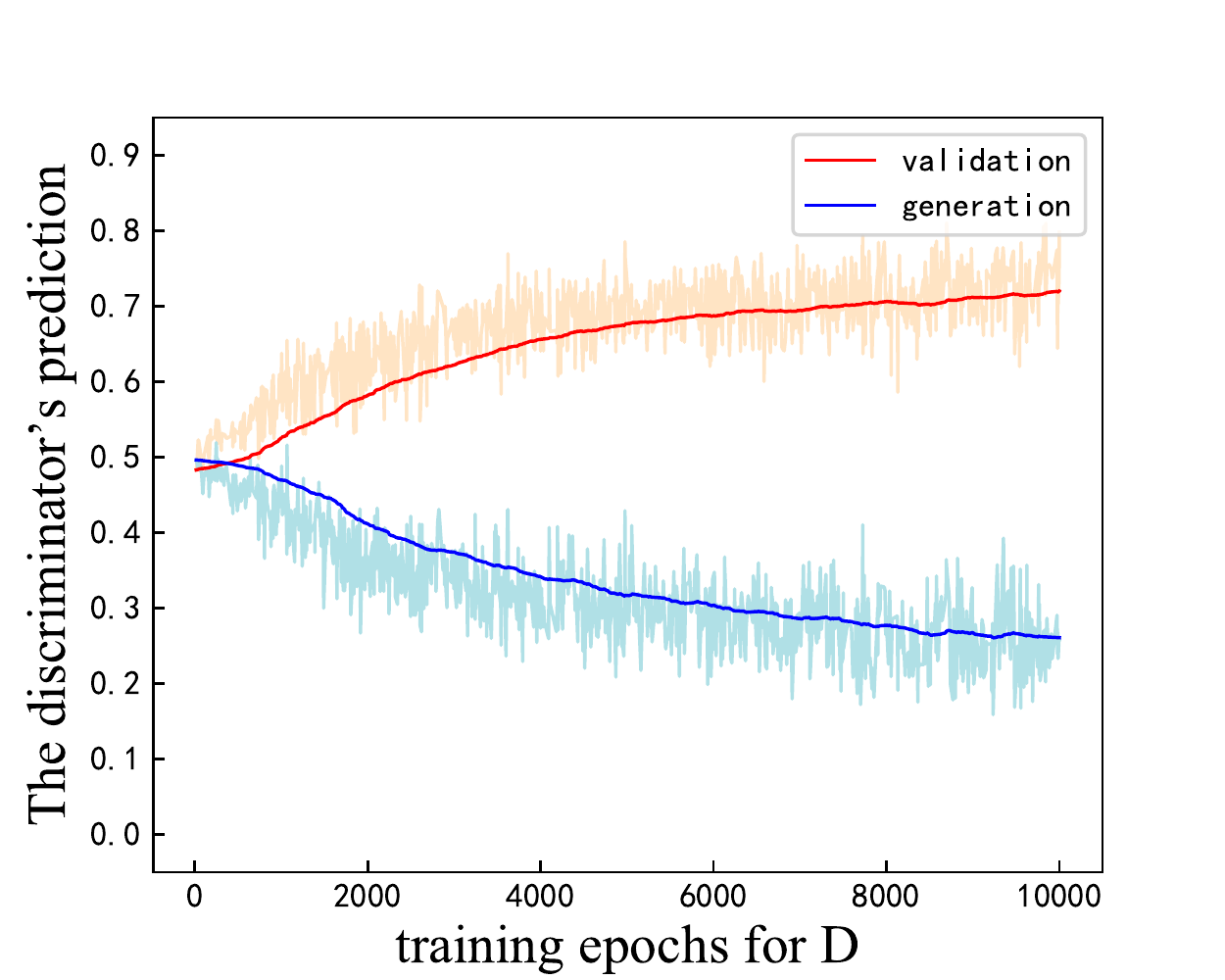}
}

\caption{The results of pre-train SeqGAN' generator 80 epochs on EMNLP dataset. All the pale lines denote batch instances' discrepancy and the curve is the exponential moving average on this sampled batch for each epoch.}
\label{figPretrainSeqEMNLP}
\end{figure*}

We estimate the distributional difference caused by the MLE-based generators. We first train the generator for $N$ epochs and then train $D$ until it converges (it is trained for 10 thousand epochs).  For example, following \citep{Nie2019ICLR}, we train $G$ for 150 epochs, and at that time, the PPL is the smallest value measured by the validation set. Then, $D_\phi$ is trained following the procedure in section 3. Figure \ref{figPretrainSeqEMNLP} shows the results on the EMNLP dataset. From this figure, we can see that (1) $D_\phi$ convergences after about 5,000 epochs. (2) There is always $u_d + u_\theta \approx 1$ everywhere. (3) When it convergences, the $u_d$ is 0.7 and $u_\theta$ is 0.3. More results are illustrated in appendix B.1.

Considering the smoothed value on one batch rather than the prediction on the whole data, we use the convergence discriminator to predict on the all validation data and generated data \footnote{According to section 3, we sample generated instances as much as test instances.}. Table~\ref{Tb:pre-train} summaries the discrepancy across two models and two datasets. It shows that the difference between real text and generated text is huge.  

\begin{table}[h]
  \caption{The discrepancy across two models and two datasets in pre-training. For both $d_a$ and $d_s$, lower is better.}
   \label{Tb:pre-train}
   \begin{center}
   \begin{tabular}{|c|c|c|c|c|}
   \hline
   Model & Dataset  &  $\hat{d}_s$ & $\hat{d}_a$ & Accuracy \\
   \hline
   \multirow{2}{*}{SeqGAN} & COCO & 0.42 & 0.44  & 0.71 \\
   & EMNLP & 0.57 & 0.47  & 0.78 \\
   \hline
   \multirow{2}{*}{RelGAN} & COCO & 0.64 & 0.45  & 0.82 \\
   & EMNLP & 0.52 & 0.31  & 0.76 \\
 
 \hline
 \end{tabular}
 \end{center}
\end{table}

\subsection{Could the detected discrepancy by $\hat{D}_\phi$ improve the generator?}
In this section, we will explore whether the above discrepancy detected by $\hat{D}_\phi$ in pre-training could improve $G$. It should be noted that the $\hat{D}_\phi$ is well pre-trained. We select out the best pre-train epochs for $G$. It is updated according to the signals from the $\hat{D}_\phi$. In order to verify the effect of those feedback signals, we generate many instances rather only several batch-size ones are used to adjust the generator's parameters. 

Then, fixing $G$, we re-train $D$ with 10,000 epochs to get a new convergence discriminator, named $\hat{D}_\phi'$, for computing two distributional functions according to equation \ref{equ6} and \ref{equ8}. Unfortunately, in the view of absolute discrepancy or approximated discrepancy, the discrepancy always overpass the original value which is computed in pre-training. This demonstrates that the generator is not improved yet. Figure \ref{figE2} illustrates a comparison. More experimental results are shown in appendix B.2.

\begin{figure*}[htbp]
\centering
\subfigure{
\includegraphics[width=0.8\columnwidth]{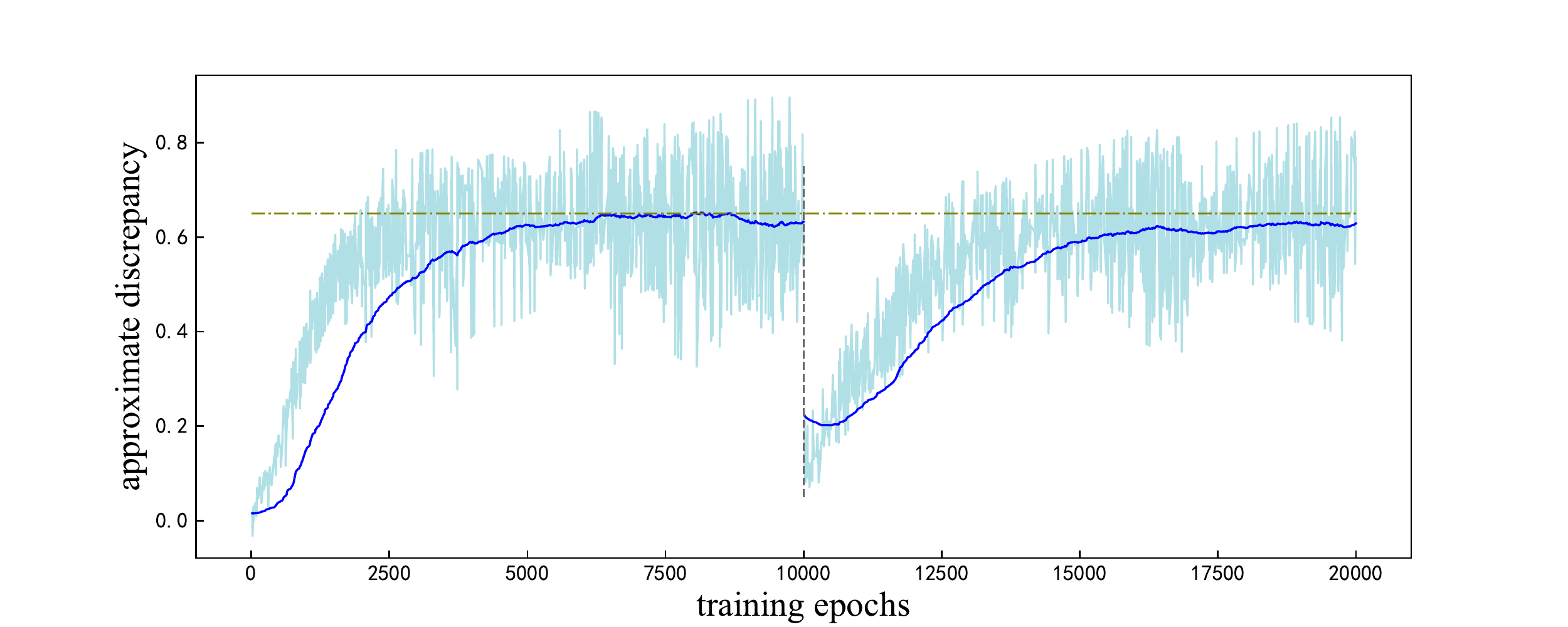}
}
\quad
\subfigure{
\includegraphics[width=0.8\columnwidth]{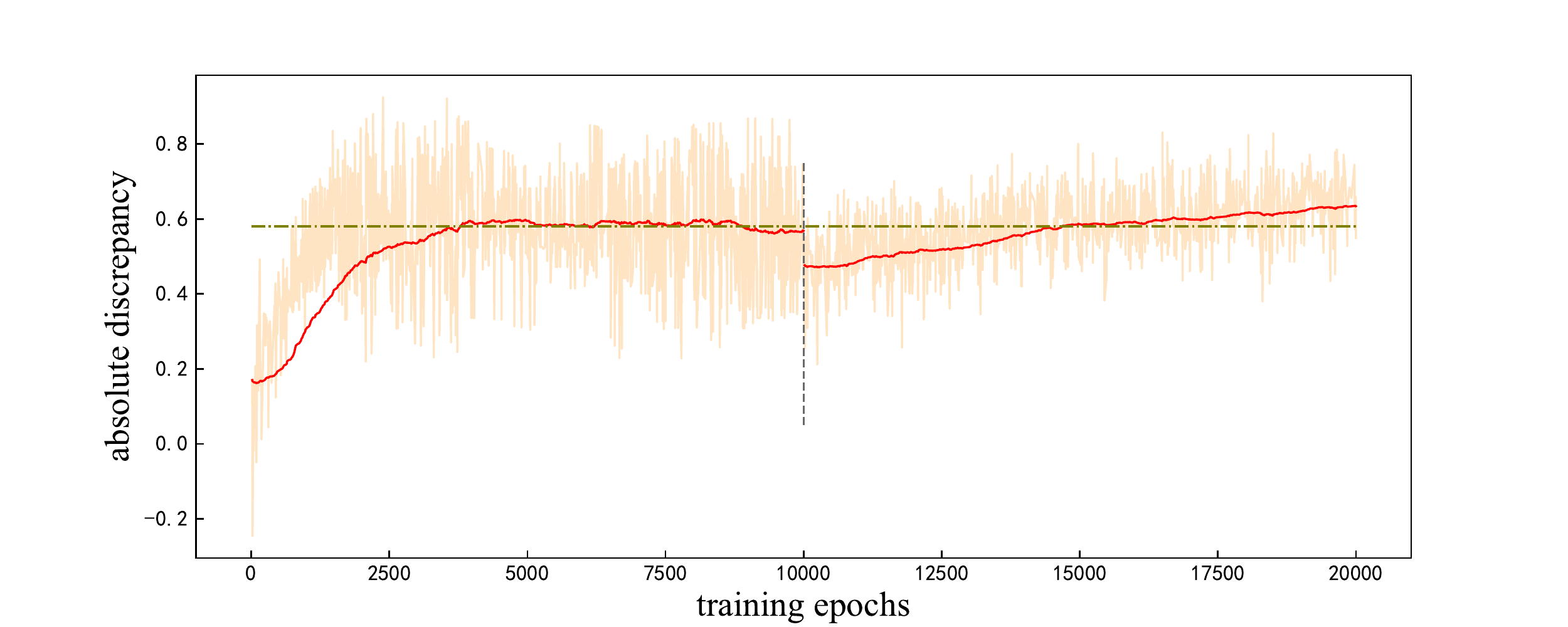}
}

\caption{The comparison of discrepancy between pre-train and the generator is updated with the feedback signals from $\hat{D}_\phi$ which is obtained from pre-train. The vertical dash line
represents the end of pre-training. SeqGAN's generator is pre-trained 80 epochs and the dataset is COCO.}
\label{figE2}
\end{figure*}

Besides following \citep{Zhu2018Texygen}, We propose a new methods train $G$. Rather than all the generated instances are used to update $G$, only the ones who are assigned relative high scores by $D$ are used. We denote it as HW. The reason is that we think the higher score instances maybe more informative than the lower ones. In fact, only the relative low scores samples used to adjust the generator is also experimented. Regretfully, all of them fail. Table~\ref{Tb:beginAD} list the discrepancy across two datasets. It again demonstrated that the absolute measure is necessary. Appendix B.3 show the results of several thresholds are set for selecting out more informative generated instances. 

\begin{table}[h]
  \caption{The compare between the absolute discrepancy in pre-training and $G$ updated by $\hat{D}_\phi$'s feedback signal. \#samples denotes the amount of the generated data is used for updating the $G$. For example, 2S means the generated instances is two times as the amount of training data. Random denotes the existing way but the other row are the results according to HW. $<0.3-0.5$ means the generated instances whose score are between 0.3 and 0.5 assigned by $D$, are selected out.}
   \label{Tb:beginAD}
   \begin{center}
   \begin{tabular}{|c|c|c|c|c|c|c|c|c|c|c|}
   \hline
   Dataset  &  \multicolumn{5}{c|}{COCO} & \multicolumn{5}{c|}{EMNLP} \\
   \hline
   pre-Train & \multicolumn{5}{c|}{\textbf{0.42}}  &  \multicolumn{5}{c|} {\textbf{0.57}}  \\
   \hline
   \#samples & 0.1S & 0.5S & 1S & 2S & 5S & 0.1S & 0.5S & 1S & 2S & 5S \\
   \hline
   random & 0.52 & 0.45 &  0.54 & 0.53 & 0.55 & 0.68 & 0.68 & 0.67 & 0.67 & 0.67 \\
   \hline
   $<0.3$ & 0.60  & 0.58  & 0.54  & 0.54  & 0.72  & 0.77  & 0.76  & 0.73  & 0.76  & 0.73 \\
   \hline
   $0.3-0.5$ & 0.44 & 0.56 & 0.58 & 0.60 & 0.55 & 0.68 & 0.67 & 0.66 & 0.63 & 0.66 \\
   \hline
   $0.5-0.9$ & 0.52 & 0.51 & 0.47 & 0.54 & 0.58 & 0.66 & 0.57 & 0.60 & 0.64 & 0.62 \\
   \hline
   $\geq0.9$ & 0.68 & 0.46 & 0.50 & 0.44 & 0.58 & 0.66 & 0.60 & 0.59 & 0.60 & 0.62 \\
   \hline
   
 \end{tabular}
 \end{center}
\end{table}

\subsection{A third party discriminator evaluates these language GANs}
\begin{figure*}[h]
\centering
\subfigure[SeqGAN]{
\includegraphics[width=0.47\columnwidth]{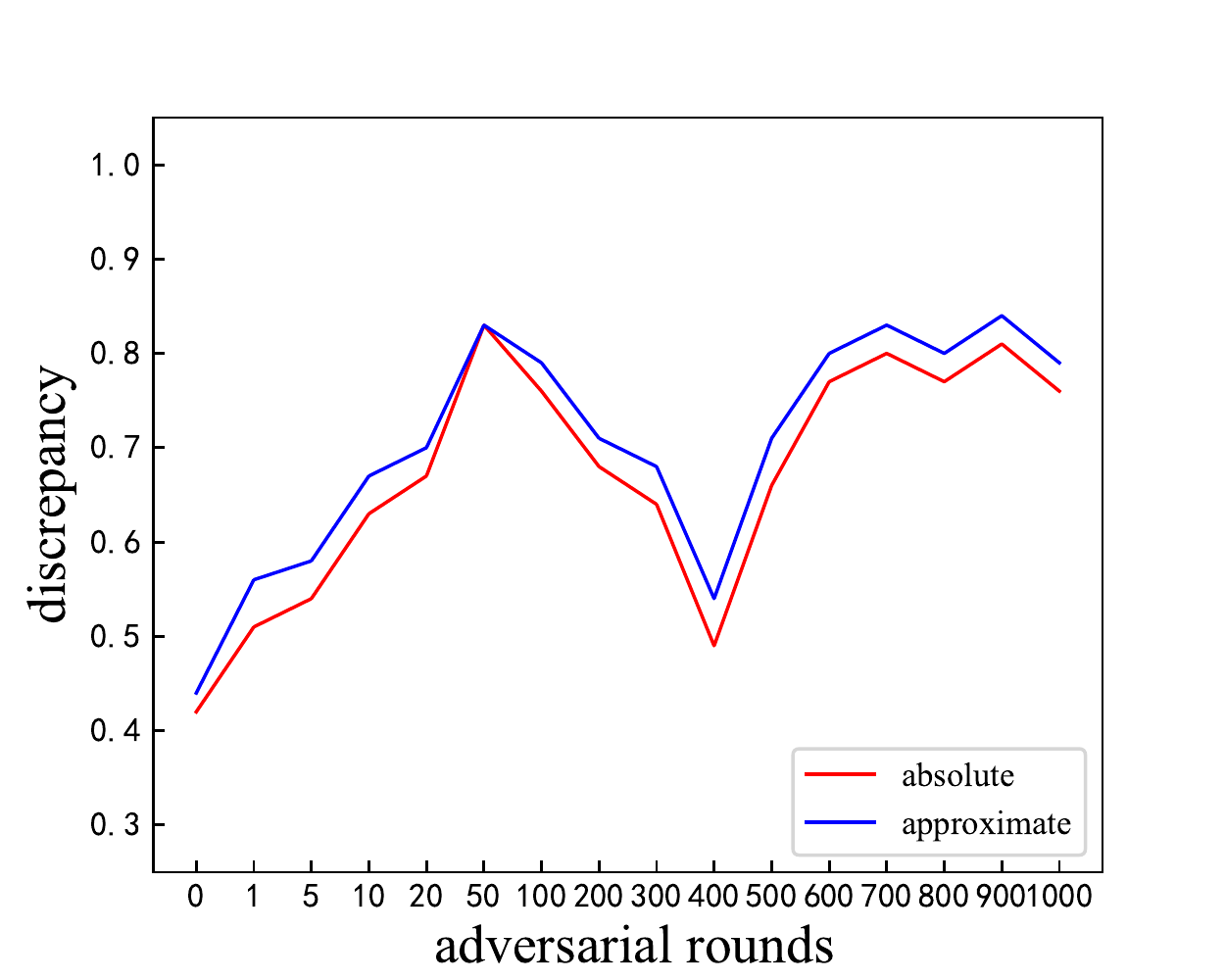}
}
\quad
\subfigure[RelGAN]{
\includegraphics[width=0.47\columnwidth]{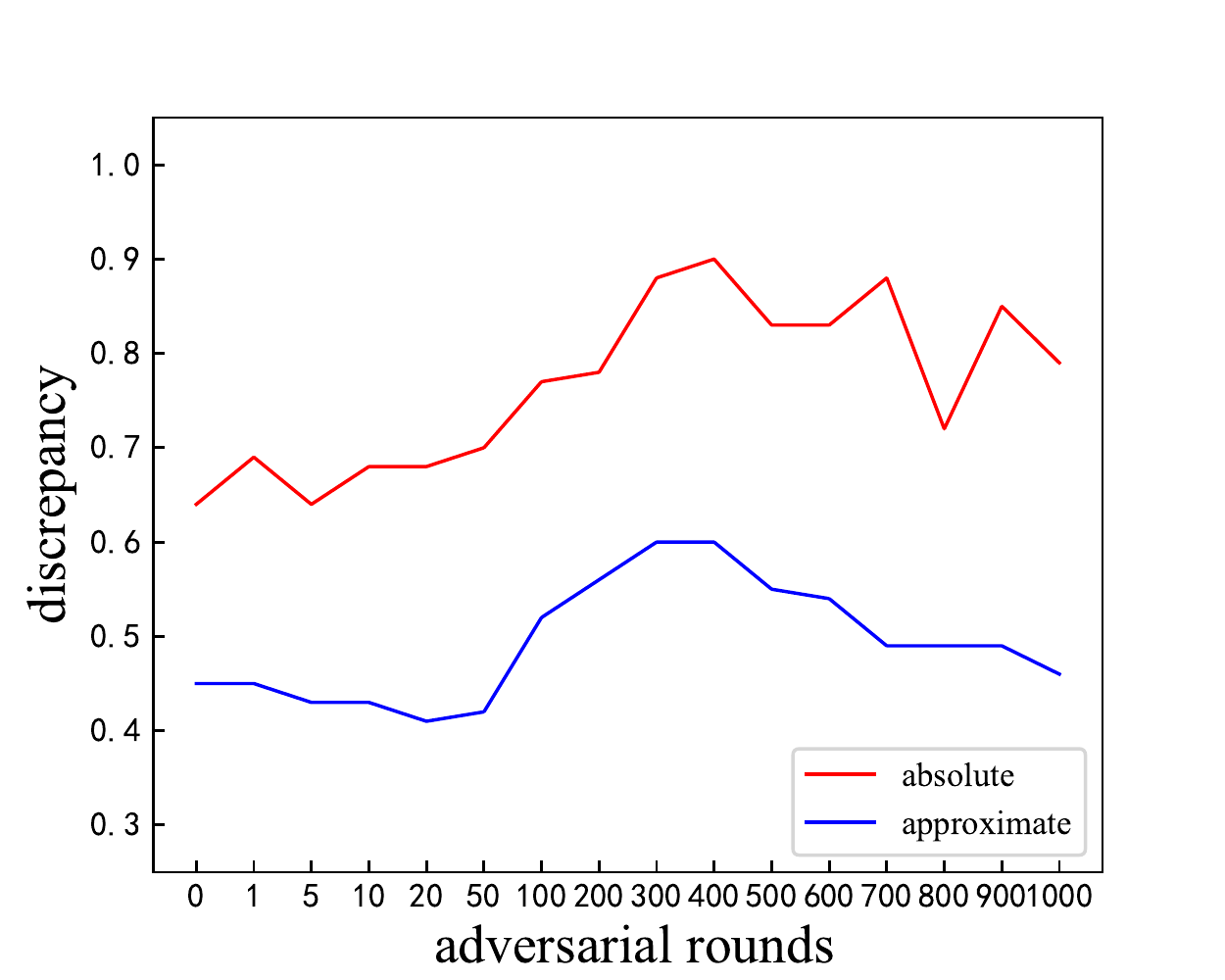}
}

\caption{A third party discriminator evaluates two GANs' performance variety on COCO. }
\label{figADV}
\end{figure*}

In order to evaluate different adversarial learning's GANs, we use a third party discriminator $D_3$ which is a clone of the discriminator in its counterpart language GAN except for the parameters' values. Given a round, we train $D_3$ many epochs (making sure it convergence) with real text and the generated text at this round. Then, two distributional functions are computed according to its prediction. Figure \ref{figADV} shows the result. In the view of both approximate discrepancy and absolute discrepancy, the difference of the distribution on real text and generated text does not decrease with more adversarial learning rounds are adapted. Once again, it shows that the approach of the existing language GANs can not improve text generation.

\section{Related Work}
Many GAN-based models are proposed to improve traditional neural language model. SeqGAN \citep{Yu2016SeqGAN} is the first one and tries by attacking the non-differential issue by resorting to RL. By applying policy gradient \citep{sutton2000policy} method, it optimizes the LSTM generator with rewards received through Monte Carlo (MC) sampling. Many researchers, such as RankGAN \citep{Lin2017Adversarial}, MailGAN \citep{Tong2017Maximum} follow this way although its in-effective in MC search. The RL-free model, for an example GSGAN \citep{Jang2016Categorical}, contains applying continuous approximating softmax function and working on latent continuous space directly. TextGAN \citep{salimans2016improved} adds Maximum Mean Discrepancy to the original objective of GAN based on feature matching. RelGAN \citep{Nie2019ICLR} is a state-of-the-art model which  uses relation memory \citep{santoro2018relational}, which allows for interactions between memory slots by using the self-attention mechanism \citep{Vaswani2017Attention}. We select out SeqGAN and RelGAN as representatives for experiment. The results show that the adversarial learning does not work for either of these models.

\cite{Caccia2019falling} first argues the current evaluation measures correlate with human judgment \citep{DBLP:journals/corr/abs-1804-07972} is treacherous. They furthermore propose temperature sweep which evaluates model at many temperature settings rather than only one. By using this metric, they find a well-adjust language model can beat those considered language GANs. \cite{Semeniuta2018On} and \cite{He2019Quantifying} also argues GAN-based models are weaker than a LM, because they observe the impact of exposure bias is not severe. \cite{He2019Quantifying} furtherly quantify the exposure bias by using conditional distribution. Neither designing a better metric nor showing the weakness of the language GANs, we try to investigate language GANs in mechanism and quantify the discrepancy between real texts and generated texts both in pre-train and adversarial learning.

\section{Conclusion and Future work}
We present two directly metric functions to measure the discrepancy between real text and generated text. It must be noted that they are independent of any text generation method including GANs-based. Numerous experiments show that this discrepancy does exist. We try some methods to update the parameters of generator according to the detected discrepancy signals. Unfortunately, the distributional difference between real data and generated data does not decrease. It is hard to improve generator with these signals. Finally, We use a third part discriminator to evaluate the effectiveness of GAN and find with more adversarial learning epochs, the discrepancy increases rather than decreasing. It shows the existing language GANs do not work at all. 

\section*{acknowledgments}
We thank Hong Yang for running codes in GPU server. We also thank Diana McCarthy for suggestions on the text. This work is supported by the National Natural Science Foundation of China (61373056).

\bibliography{iclr2020_conference}
\bibliographystyle{iclr2020_conference}

\clearpage
\appendix
\section{Formula induction}
We show the derivation of Equation \ref{equ8},

\begin{equation}\label{equA-1}
\begin{aligned}
d_s &= \frac{1}{2}\int{\big|p_d(x) - p_\theta(x)\big|dx} \\
    &= \frac{1}{2}\bigg[\int_{p_d(x)\geq p_\theta(x)}{\big(p_d(x) - p_\theta(x)\big)dx} + \int_{p_d(x)<p_\theta(x)}{\big(p_\theta(x) - p_d(x)\big)dx} \bigg]\\
    &= \frac{1}{2}\bigg[\int_{p_d(x)\geq p_\theta(x)}{p_d(x)dx} + \int_{p_d(x)<p_\theta(x)}{p_\theta(x)dx} - \int_{p_d(x)\geq p_\theta(x)}{p_\theta(x)dx} \\ 
    & \qquad\qquad\qquad\qquad\qquad\qquad\qquad\qquad\qquad\qquad\qquad - \int_{p_d(x)< p_\theta(x)}{p_d(x)dx} \bigg]\\
    &= \frac{1}{2}\bigg[\mathbb{E}_{\substack{x\sim{p_d(x)}\\ {p_d(x)\geq p_\theta(x)}}}\big(1\big)  + 
    	\mathbb{E}_{\substack{x\sim{p_\theta(x)}\\ {p_d(x)<p_\theta(x)}}}\big(1\big) -  
    	\mathbb{E}_{\substack{x\sim{p_\theta(x)}\\ {p_d(x)\geq p_\theta(x)}}}\big(1\big) - 
    	\mathbb{E}_{\substack{x\sim{p_d(x)}\\ {p_d(x)<p_\theta(x)}}}\big(1\big) \bigg]\\
    &= \frac{1}{2}\bigg[\mathbb{E}_{\substack{x\sim{p_d(x)}\\ {z\geq 0.5}}}\big(1\big)  + 
    	\mathbb{E}_{\substack{x\sim{p_\theta(x)}\\ {z<0.5}}}\big(1\big) -  
    	\mathbb{E}_{\substack{x\sim{p_\theta(x)}\\ {z\geq 0.5}}}\big(1\big) - 
    	\mathbb{E}_{\substack{x\sim{p_d(x)}\\ {z<0.5}}}\big(1\big) \bigg]\\
\end{aligned}
\end{equation}

where $z = \frac{p_{d}(x)}{p_{d}(x) + p_\theta(x)}$.

\clearpage

\section{Appendix B}

\subsection{The rest three examples in pre-train. }
\begin{figure*}[htbp]
\centering
\subfigure{
\includegraphics[width=0.47\columnwidth]{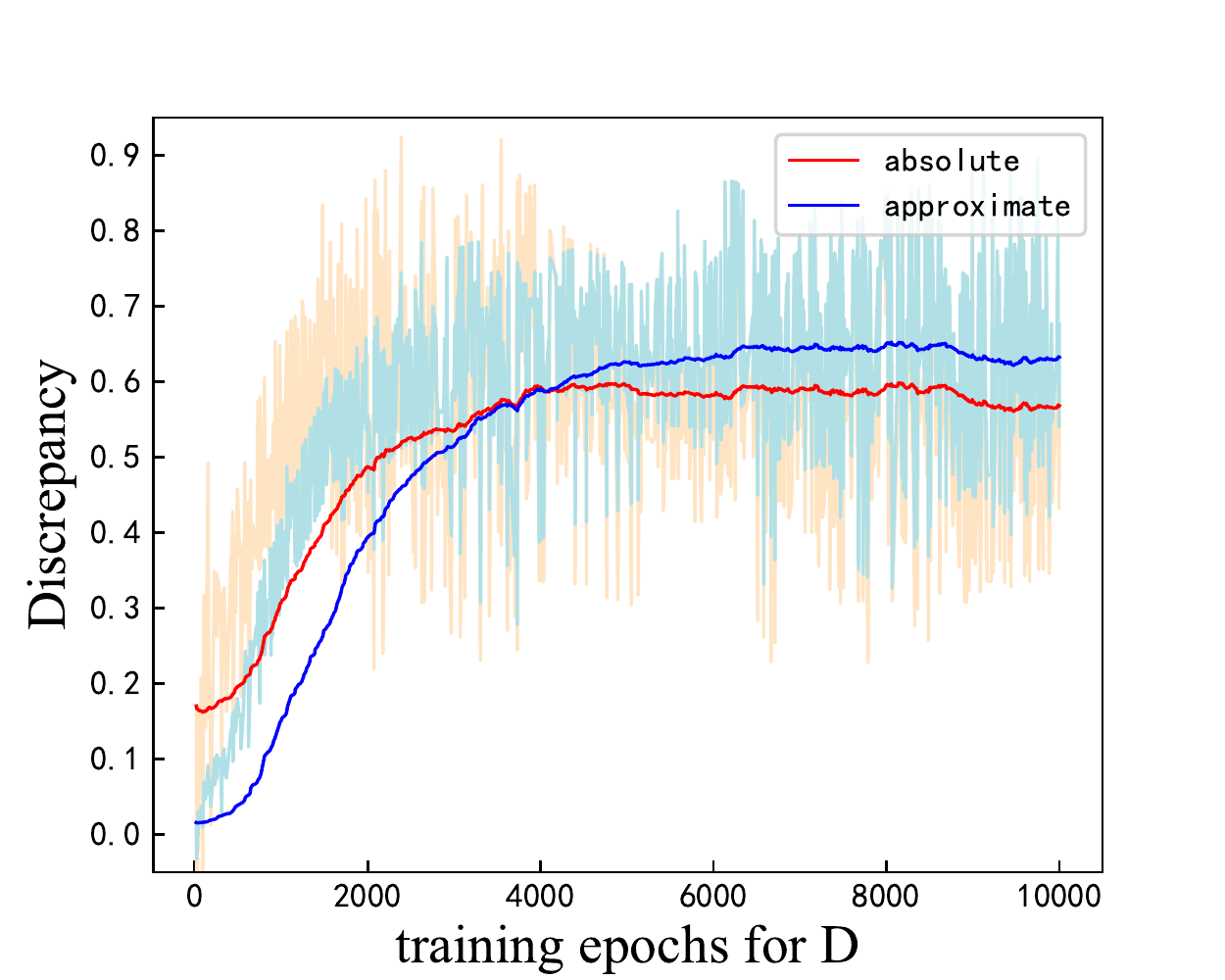}
}
\quad
\subfigure{
\includegraphics[width=0.47\columnwidth]{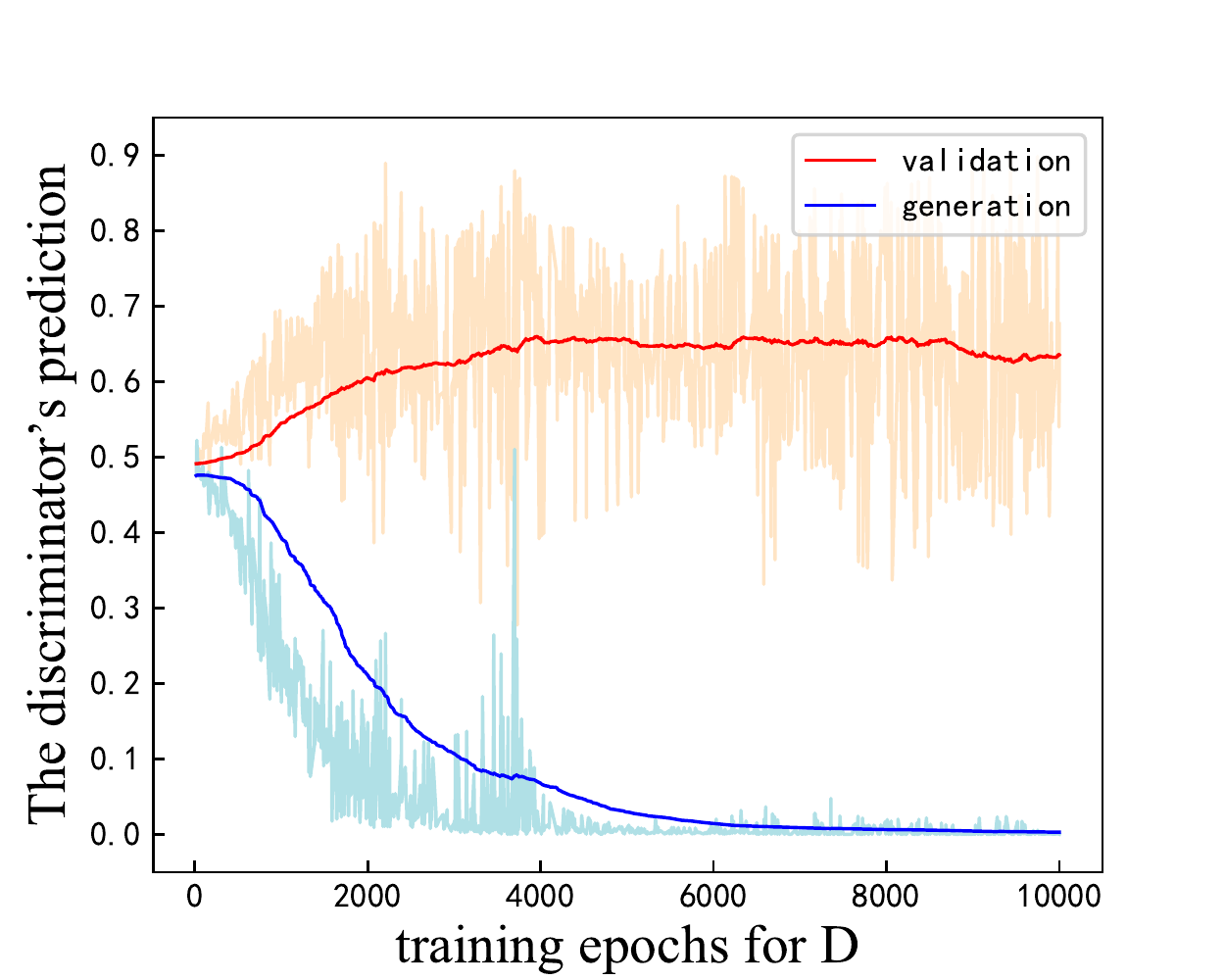}
}
\subfigure{
\includegraphics[width=0.47\columnwidth]{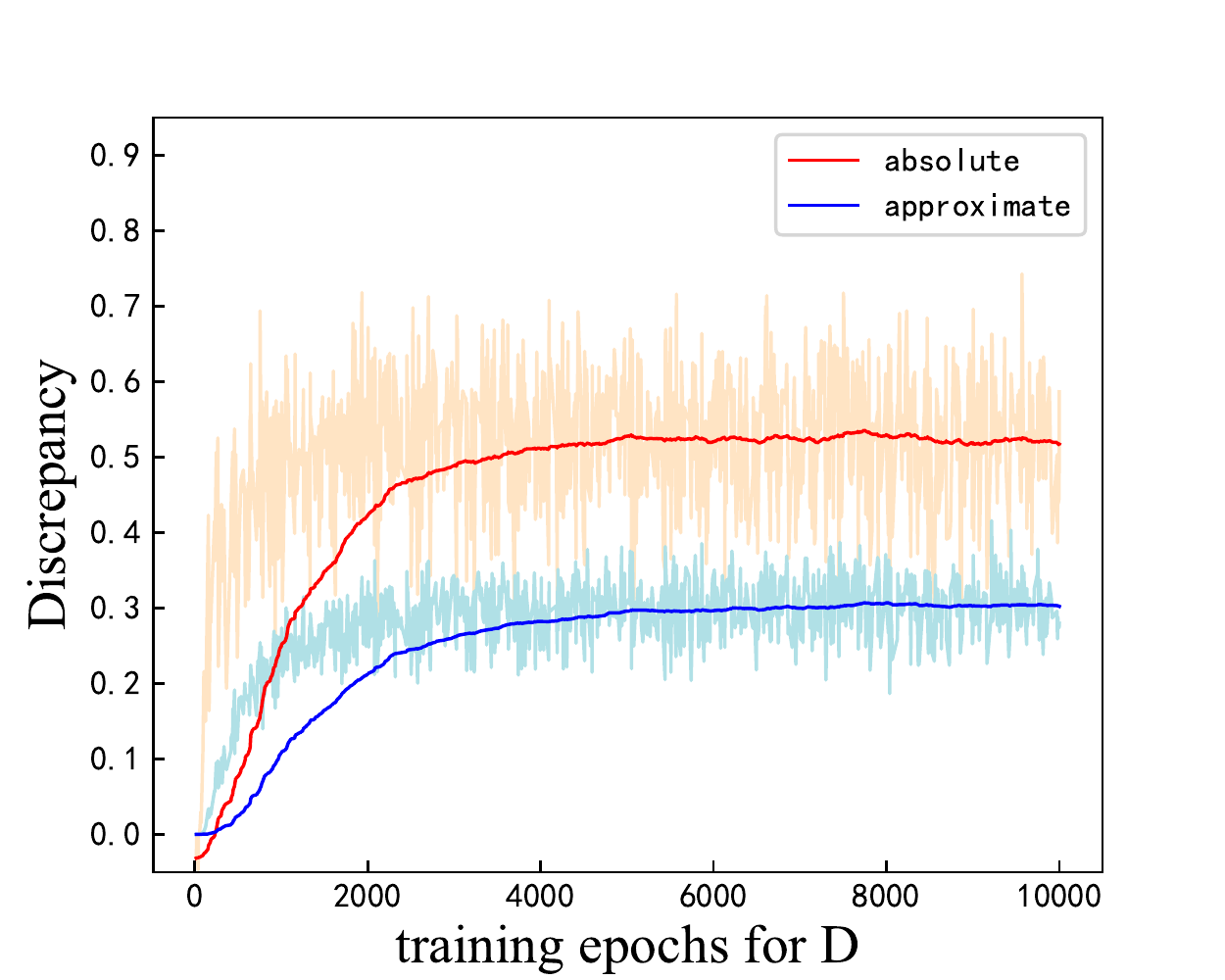}
}
\quad
\subfigure{
\includegraphics[width=0.47\columnwidth]{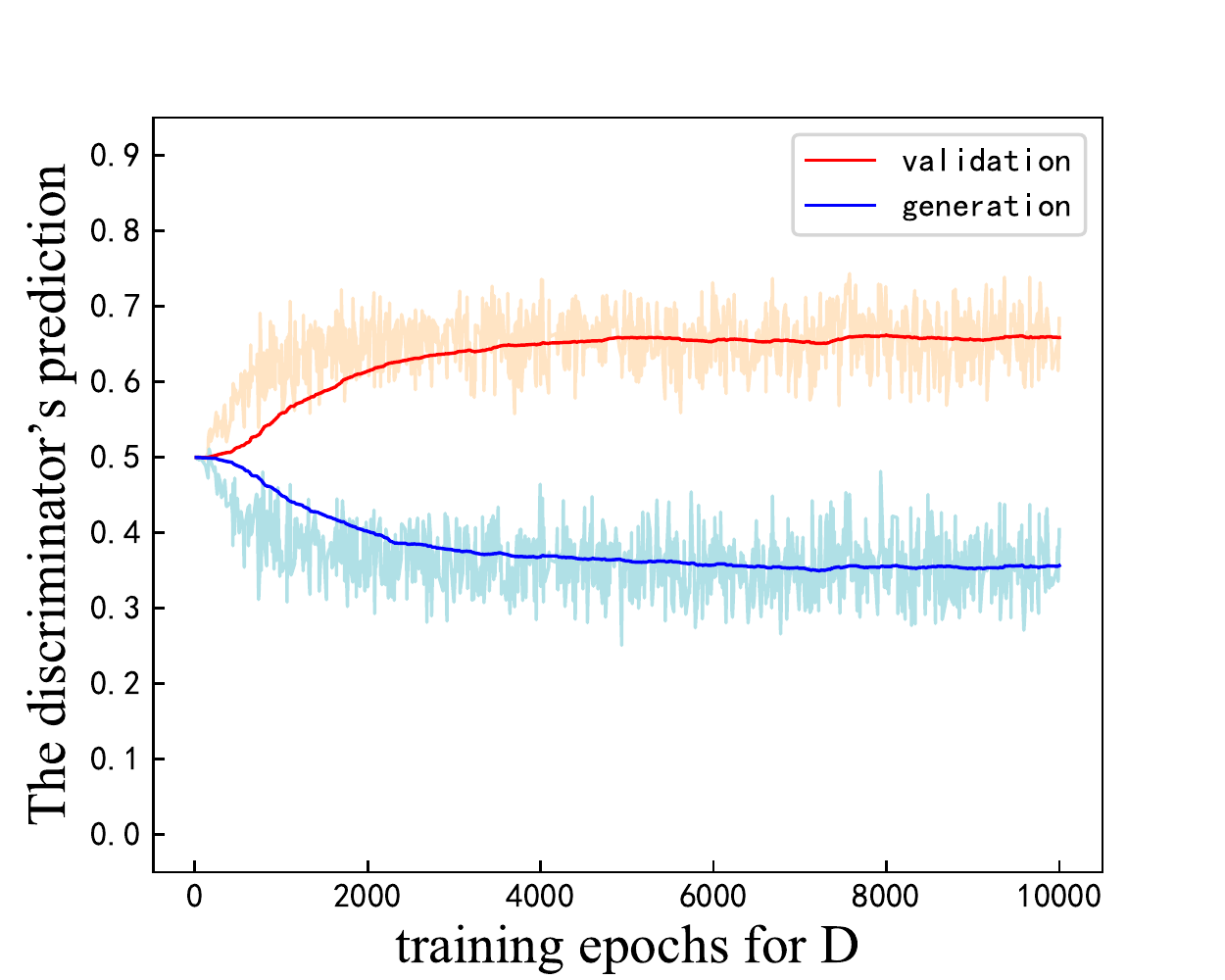}
}
\subfigure{
\includegraphics[width=0.47\columnwidth]{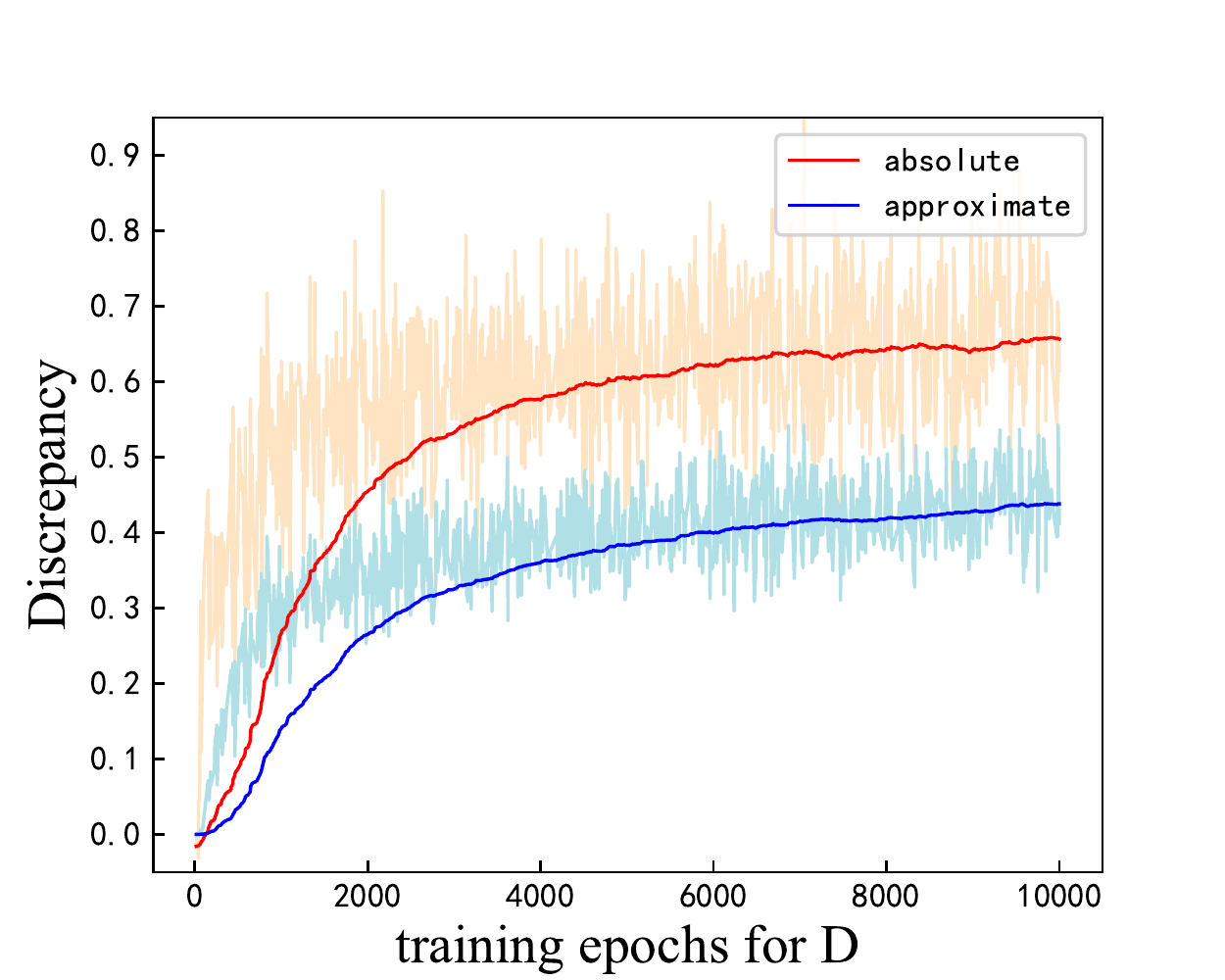}
}
\quad
\subfigure{
\includegraphics[width=0.47\columnwidth]{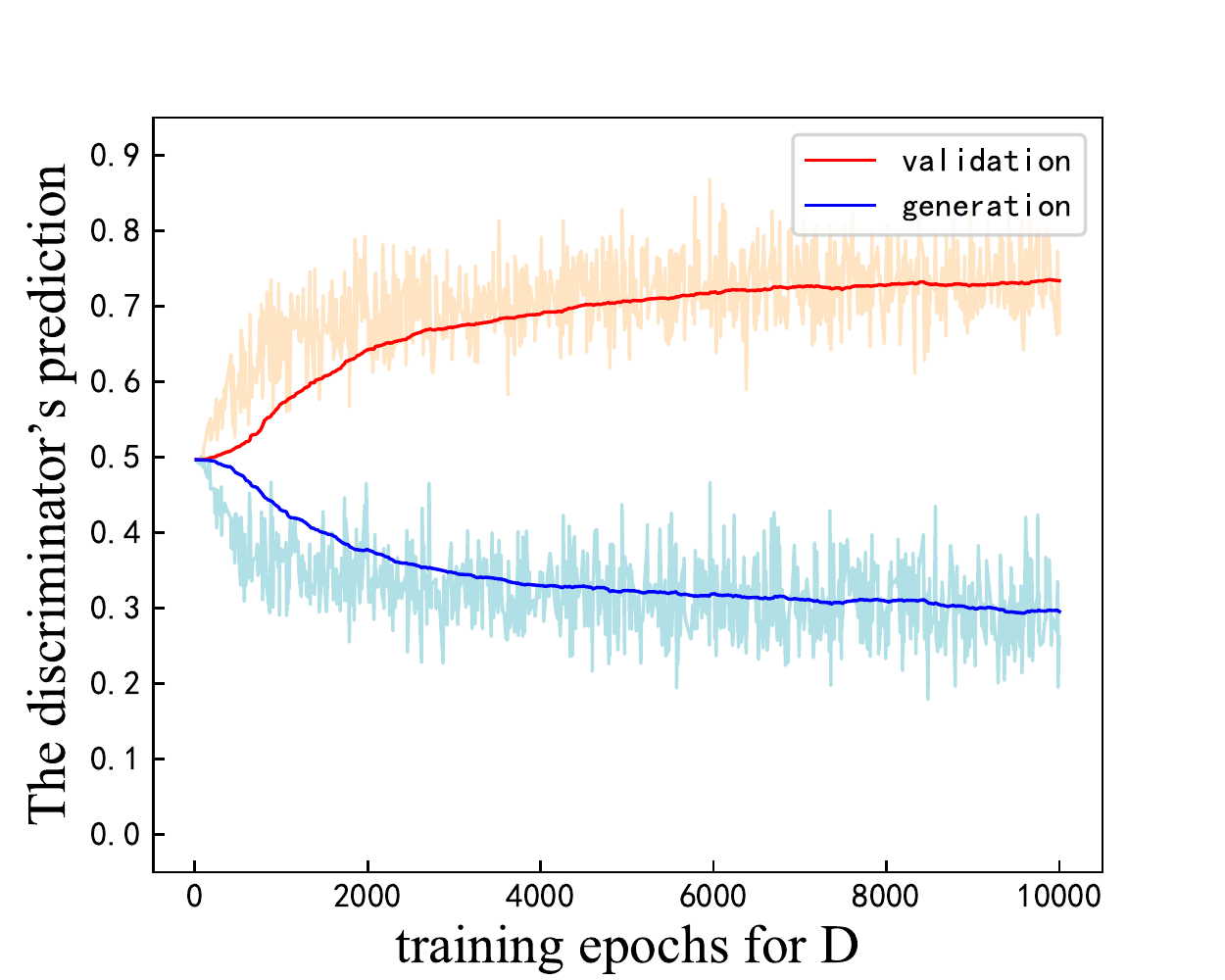}
}

\caption{The results of pre-train SeqGAN' generator 80 epochs on the COCO dataset (upper), RelGAN on EMNLP (middle) and RelGAN on COCO(below). All the pale lines denote the batch instances' discrepancy and the curve is the exponential moving average on this sampled batch for each epoch.}
\label{figPretrainSeqCOCO}
\end{figure*}
\clearpage

\subsection{The discrepancy comparison on the EMNLP dataset between pre-training and updated LM.}

\begin{figure*}[htbp]
\centering
\subfigure{
\includegraphics[width=1.0\columnwidth]{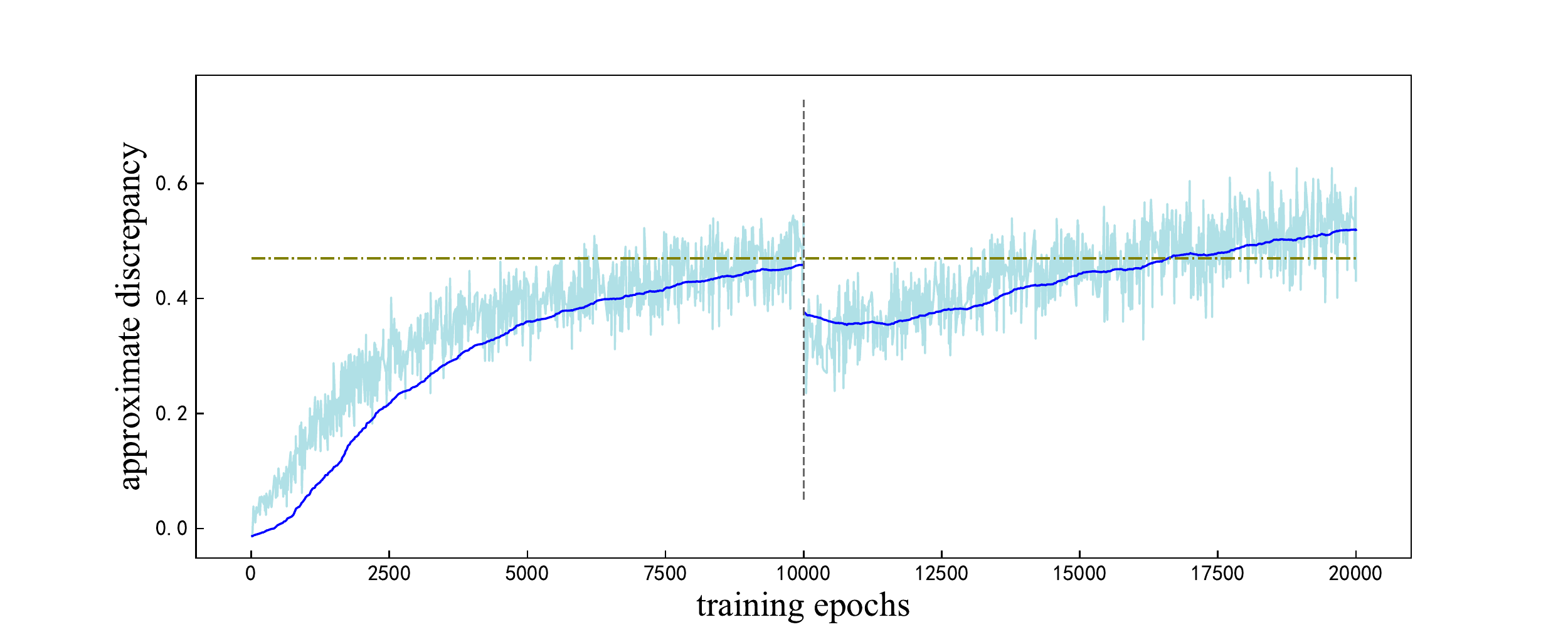}
}
\quad
\subfigure{
\includegraphics[width=1.0\columnwidth]{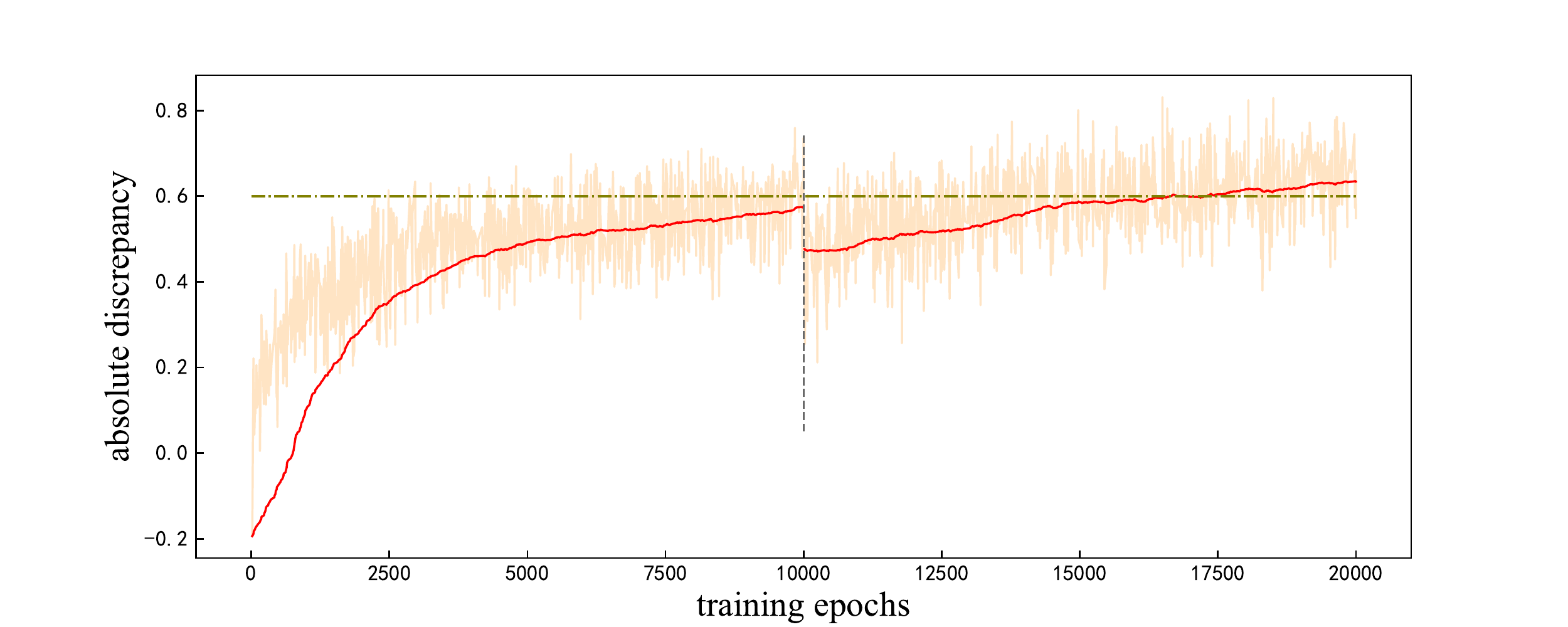}
}

\caption{The comparison of discrepancy between pre-train and the generator is updated with the feedback signals from $\hat{D}_\phi$ which is obtained from pre-train. The vertical dash line
represents the end of pre-training. SeqGAN' generator is pre-trained 80 epochs and the dataset is EMNLP.}
\label{figAPPB2}
\end{figure*}

\subsection{The comparison of the approximate discrepancy between random and HW.}

\begin{table}[h]
  \caption{The comparison between the approximate discrepancy in pre-training and $G$ is updated by $\hat{D}_\phi$'s feedback signal. \#samples denotes the amount of the generated data is used for updating the $G$. For example, 2S means the generated instances is two time as the amount of test data. Random denotes the existing way but the other row are the results according to HW. $<0.3-0.5$ means the generated instances whose score are between 0.3 and 0.5 assigned by $D$, are selected out.}
   \label{Tb:beginAD}
   \begin{center}
   \begin{tabular}{|c|c|c|c|c|c|c|c|c|c|c|}
   \hline
   Dataset  &  \multicolumn{5}{c|}{COCO} & \multicolumn{5}{c|}{EMNLP} \\
   \hline
   pre-Train & \multicolumn{5}{c|}{\textbf{0.44}}  &  \multicolumn{5}{c|} {0.47}  \\
   \hline
   \#samples & 0.1S & 0.5S & 1S & 2S & 5S & 0.1S & 0.5S & 1S & 2S & 5S \\
   \hline
   random & 0.57 & 0.50 &  0.58 & 0.58 & 0.60 & 0.55 & 0.52 & 0.50 & 0.52 & 0.53 \\
   \hline
   $<0.3$ & 0.65  & 0.62  & 0.59  & 0.60  & 0.75  & 0.64  & 0.62  & 0.61  & 0.61  & 0.61 \\
   \hline
   $0.3-0.5$ & 0.49 & 0.60 & 0.62 & 0.64 & 0.56 & 0.51 & 0.50 & 0.49 & 0.49 & 0.49 \\
   \hline
   $0.5-0.9$ & 0.57 & 0.55 & 0.51 & 0.58 & 0.62 & 0.49 & 0.46 & 0.45 & 0.45 & \textbf{0.44} \\
   \hline
   $\geq0.9$ & 0.55 & 0.55 & 0.50 & 0.61 & 0.47 & 0.47 & 0.45 & 0.45 & 0.45 & 0.46 \\
   \hline
   
 \end{tabular}
 \end{center}
\end{table}

\clearpage

\section{Generated sentences on COCO image captions dataset}

\begin{table}[h]
  \caption{Generated sentences who are scored 0.9 or higher by  $\hat{D}_\phi$ at the end of pre-training. Obviously, they are better than the next sentences listed in table~\ref{Tb:generatedLow}}.
   \label{Tb:generatedHigh}
   \begin{center}
   \begin{tabular}{l}
   \hline
    the ground and sink are under the window of a pink covered pot . \\
a toilet sitting next to a toilet next to a green organized kitchen . \\
the bathroom contains a mop that rolls of flowers next to the vanity do two sinks .  \\
a vehicle trailer meeting a corner of a street .  \\
a small dog is sitting in a purse with text .  \\
two men are riding an outside of the window of a street  \\
a porcelain tea pot by a window  \\
a humongous jumbo jet with chain chair on the ground .  \\
large white bus some luggage onto one the other smiling no front tire .  \\
a wooden ship with rusted heater flying high in a sky .  \\
the side outside with a giraffe doors  \\
the bathroom has a pedestal table with pots on it  \\
bathroom with wooden cabinets and a washer , curtain and bottles growing on the flooring .  \\
the side of a person looks at a woman on the back seat .  \\
a bottle of whiskey in the bathroom with broken toilet   \\
a ramp extends to the top of an air mattress .   \\
a photograph of a palm wall in the glass 's lap top .   \\
a smiley face stand with 4 per gallon .   \\
several bicycles parked stuffed in a display using it .   \\
two dogs on top of a car at an amusement table .   \\
a white airplane flying above someone with his friend parked behind it terminal .   \\
a bathroom water from a shower with cobbler and pink tile walls .   \\
a toilet is bathroom with multiple monitors paper .   \\
a red fire hydrant displaying the woman in a park and many headlights .   \\
a tiled bathroom with a claw view of toilet and soap dispenser   \\
a group of ripe bananas in their hands on the back seat .   \\
a motorcycle parked next to a stone building next to the road .   \\
an airplane flies low to people aboard a boat with a cemetery a frisbee .   \\
a bath room with a toilet behind it   \\
a homemade cake in home bathroom has a sink , a , shower , and a window .   \\
a truck is shown as the parking lot corner .   \\
a bathroom is lit with foods such as : to a web job graze .   \\
porcelain toilet in a blue toilet seat .   \\
a simple white bathroom with a large white fridge , and shower .   \\
a large delta passenger airplane flying through a cloudy sky .   \\
a bathroom with a mirror , dishwasher , dishwasher , and tub/shower .   \\
a blue plane is walking as seen all towards the usual number of tonic .   \\
two adorable chubby dogs in a group of pots   \\
four airplanes sitting on top of a building in the blue sky .   \\
horse is racing on his motorcycle while maintenance are walking next to the river .  \\
an image of a old bathroom with a bottle of spirits suggest tuscan decor , laptop .  \\
an outdoor art bus is facing toward another snowboarder  \\
a kitchen filled with in tables and two computer monitors . \\ 
this is an image of a motorcycle parked down by a bench .  \\
a crowd of people sitting on a bench next to street  \\
an electronic cat in a mens restroom next to multiple dark and pans along under a screen table .  \\
two woman in her phone next to a vehicle .  \\
a person riding a motor moped at a playground .  \\
two dirty dog standing in front of a car .  \\
a white bathtub sits next to a mirror in a small closed . \\
    
   \hline
   
 \end{tabular}
 \end{center}
\end{table}

\begin{table}[h]
  \caption{Generated sentences who are scored 0.1 or lower by  $\hat{D}_\phi$ at the end of pre-training. Obviously, they are lower quality than the sentences listed in the above table.}
   \label{Tb:generatedLow}
   \begin{center}
   \begin{tabular}{l}
   \hline
    a white metal structure with a backpack on top .  \\
a bath tub next to the toilet in middle of a nice plate .  \\
rams lights and need onto the outside of a zoo .  \\
an airplane parked on the ground in front of a building .  \\
a white vw car passing in front of a car .  \\
an airplane flying over an airport terminal .  \\
two people riding bikes on an outdoor from police officer .  \\
an old propeller airplane parked off .  \\
a man standing down by side on a street .  \\
a image of a toilet , mirror and painting on the wall  \\
two toilets walking a lipstick , a television .  \\
two doves sit on a bathroom counter with wooden cabinets and a bucket .  \\
this enclosure has a white toilet next to the sink .  \\
the dining area is wide two people riding her back to carve above them .  \\
two men stand with all line of street using indoors .  \\
a boy wearing travel and his motorcycle outside a town outside .  \\
a person brushing her teeth with her hair holding a stove .  \\
a white bathtub sitting in a bathroom with no cabinets .  \\
a bath room with two counter preparing food .  \\
a man holding up a market stand by the ground .  \\
a kitchen has large round glass serving and cabinets .  \\
a woman standing outside of a black car .  \\
a bathroom has an island in the middle granite glass .  \\
a large passenger jet flying over an airport next to the street .  \\
a tiger cat sitting on top of a window looking very clean .  \\
bicyclists holding a flip phone next to the aircraft .  \\
a walk opened in the bathroom with a jungle theme .  \\
the sink has white appliances with a child .  \\
a kitchen with a chrome toilet next to bathtub . \\ 
man standing in a blue park looking sidewalk next to a brown horse against a group of electrical boxes .  \\
an old parked motorcycle with its kickstand down to .  \\
a messy bathroom with a tub , in the sink and the mirror with no privacy .  \\
a an open kitchen tucked in a public kitchen . \\
a white brown kitchen with bowls on a counter top counter .  \\
a woman standing in red liquid under a small water fountain .  \\
a view above the toilet roll a sink with a stove .  \\
a couple of chefs up in a kitchen preparing food .  \\
a bathroom with a toilet , and mirrors from its reflection in a large sink .  \\
a bath room with a tub and refrigerator .  \\
a group of traffic light with people skiing in the mist  \\
two dogs are huddled together a screen corner on a wall .  \\
wild animals grazing in the center of a blue sky .  \\
a woman holds a spoon by off a road .  \\
a kitchen with toiletries in the just darts .  \\
motor sandwich and ride to turn across a road .  \\
a man crossing a traffic in an oriental city  \\
women turning her phone food prepares food .  \\
two stuffed animals are laptop , dishwasher , and a sink .  \\
a toilet in front of a window in a white room .  \\
large woman , on snowboards at an open purse  \\
an image of a sink , trash can .  \\
    
   \hline
   
 \end{tabular}
 \end{center}
\end{table}
\bibliographystyle{plainnat}
\end{document}